\documentclass[10pt]{article}
\usepackage{main}
\usepackage{xspace}

\makeatletter
\DeclareRobustCommand{\mmopd}{\begingroup
  \ifdim\f@size pt>12pt
    \mbox{\fontsize{15}{16}\selectfont\ttfamily\bfseries
      \color{HeadingColor}MM-OPD}\else
    \mbox{\ttfamily\bfseries\color{HeadingColor}MM-OPD}\fi
  \endgroup\xspace}
\makeatother

\papertitle{\mmopd}{Towards One More Bottleneck Between Perception and Reasoning}
\paperauthors{Jintao Tong$^{1,2}$, Yujing Lou$^{2,\ddagger}$, Zhanming Shen$^{2,3}$, Jiaqi Gu$^{2}$, Lubin Fan$^{2,\dagger}$\\Ruixuan Li$^{1,\dagger}$, Yue Wu$^{2}$, Jieping Ye$^{2}$, Yixiong Zou$^{1,\dagger}$}
\paperaffiliations{$^1$Huazhong University of Science and Technology\quad $^2$Alibaba Token Hub, Alibaba Group\\$^3$Zhejiang University\quad $^\dagger$Corresponding authors\quad $^\ddagger$Project leader}
\paperlinks{
  \paperlink{\githubicon}{Code}{https://github.com/TungChintao/MM-OPD}
  \quad
  \paperlink{\dataicon}{MM-OPD-4B}{https://huggingface.co/JosephTong/MM-OPD-4B}
  \quad
  \paperlink{\dataicon}{MM-OPD-9B}{https://huggingface.co/JosephTong/MM-OPD-9B}
  \quad
  \paperlink{\dataicon}{MM-OPD-32K}{https://huggingface.co/datasets/JosephTong/MM-OPD-32K}
}

\begin{document}

\papermaketitle

\begin{paperabstract}
Recent multimodal large language models (MLLMs) advance visual reasoning by strengthening both perception and reasoning, implicitly assuming a process that transitions seamlessly from perception to reasoning. However, we observe a counterintuitive phenomenon that challenges this assumption: holding the model, question, and decoding fixed, we replace images with their caption or code representations (\emph{symbolic views}), which seems to be redundant given the clear image structures, but the performance surprisingly improves by $10.2\%$ to $23.6\%$ across model scales and datasets.
We term this performance gap as the \textbf{\emph{Symbolic Visual Gap}} and then take a closer look at it.
Through experiments, we find that although the visual evidence can already appear in the reasoning trace for the image-input model, the symbolic-view-input model shows much higher attention to the correct evidence than the image-input model.
This suggests that despite good capabilities from current works in perception and reasoning themselves, another bottleneck exists \textbf{between} perception and reasoning in selecting perceived visual information as appropriate evidence for subsequent reasoning.
To handle this bottleneck, since the symbolic view steers attention toward correct evidence and is readily obtained at scale, it provides supervision for evidence selection without manually labeled evidence. Building on this, we introduce \mmopd, a multimodal on-policy self-distillation framework for \emph{symbolic-to-visual correction} that transfers guidance from symbolic-conditioned behavior to the image-conditioned policy through residual token-level targets, steering the model toward correct visual evidence. Experiments across benchmarks and model scales show that \mmopd improves a broad range of multimodal abilities rather than a single targeted capability, with gains in visual perception, chart and document understanding, mathematical reasoning, and general VQA.
\end{paperabstract}

\section{Introduction}
\label{sec:introduction}

Recent advances in multimodal large language models (MLLMs) have shown promising capabilities in visual reasoning.
Some efforts emphasize access to fine-grained visual information~\citep{liu2025seg,su2026pixel}, while others strengthen multi-step mathematical and logical reasoning through post-training~\citep{shao2024deepseekmath,hong2025glm},
assuming an ideal visual reasoning process that seamlessly transitions from perception to reasoning.

\begin{figure}[t]
  \centering
  \begin{minipage}[t]{0.58\linewidth}
    \vspace{0pt}
    \centering
    \includegraphics[width=\linewidth]{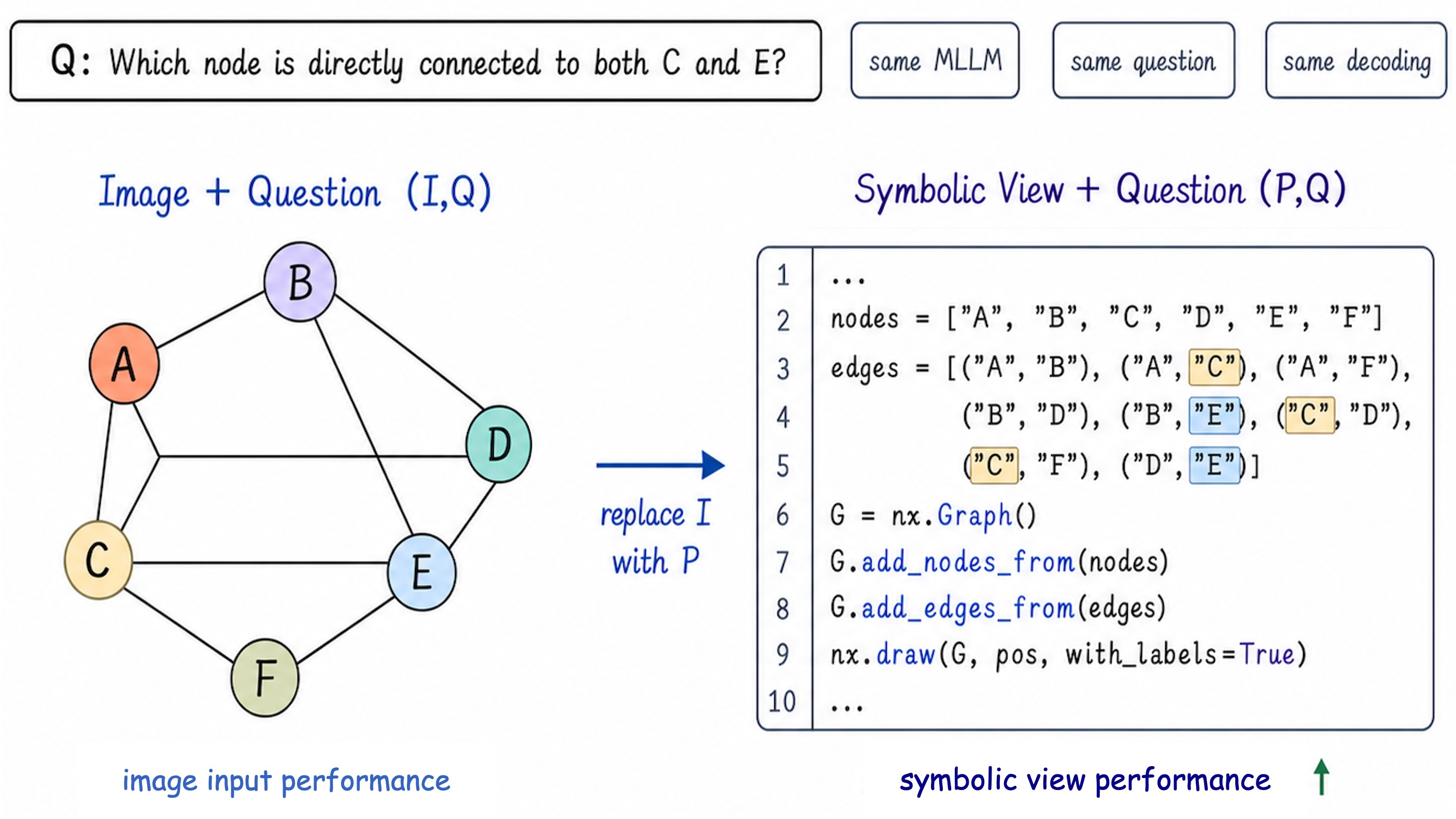}
  \end{minipage}\hfill
  \begin{minipage}[t]{0.40\linewidth}
    \vspace{0pt}
    \centering
    \resizebox{\linewidth}{!}{\begin{tikzpicture}[x=1cm, y=1cm]
      \path[fill=black!13, draw=black!45, line width=0.4pt] (9.00,5.90) circle (0.075);
      \node[anchor=west, font=\scriptsize, text=PaperInk] at (9.13,5.90) {Visual input};
      \path[fill=MidAccentColor!85!white, draw=DarkAccentColor, line width=0.45pt] (11.60,5.90) circle (0.075);
      \node[anchor=west, font=\scriptsize, text=PaperInk] at (11.73,5.90) {Symbolic input};

      \foreach \xx in {8.30,9.10,9.90,10.70,11.50,12.30,13.10,13.90} {
        \draw[black!6, line width=0.3pt] (\xx,0.90) -- (\xx,5.55);
      }
      \draw[black!50, line width=0.45pt] (8.30,0.90) -- (13.90,0.90);
      \foreach \xx/\lab in {8.30/30,9.10/40,9.90/50,10.70/60,11.50/70,12.30/80,13.10/90,13.90/100} {
        \draw[black!50, line width=0.4pt] (\xx,0.90) -- (\xx,0.84);
        \node[font=\tiny, text=PaperMuted] at (\xx,0.70) {\lab};
      }
      \node[font=\tiny\bfseries, text=PaperMuted] at (11.10,0.42) {Accuracy (\%)};

      \draw[black!10, line width=0.4pt] (7.42,4.18) -- (14.40,4.18);
      \draw[black!10, line width=0.4pt] (7.42,2.64) -- (14.40,2.64);

      \node[anchor=west, font=\scriptsize\bfseries, text=PaperInk] at (7.42,5.48) {Chart};
      \node[anchor=east, font=\tiny, text=PaperMuted] at (13.90,5.48) {symbolic view: code};
      \node[anchor=west, font=\scriptsize\bfseries, text=PaperInk] at (7.42,3.94) {Math};
      \node[anchor=east, font=\tiny, text=PaperMuted] at (13.90,3.94) {symbolic view: code};
      \node[anchor=west, font=\scriptsize\bfseries, text=PaperInk] at (7.42,2.40) {General VQA};
      \node[anchor=east, font=\tiny, text=PaperMuted] at (13.90,2.40) {symbolic view: caption};

      \node[anchor=east, font=\tiny\bfseries, text=PaperMuted] at (8.16,5.04) {4B};
      \draw[-{Stealth[length=3.2pt]}, black!32, line width=0.65pt] (11.21,5.04) -- (12.79,5.04);
      \path[fill=black!13, draw=black!45, line width=0.5pt] (11.21,5.04) circle (0.088);
      \path[fill=MidAccentColor!85!white, draw=DarkAccentColor, line width=0.55pt] (12.94,5.04) circle (0.098);
      \node[anchor=east, font=\tiny\bfseries, text=black!55] at (11.05,5.04) {66.4};
      \node[anchor=west, font=\tiny\bfseries, text=DarkAccentColor] at (13.10,5.04) {88.0};
      \node[font=\tiny\bfseries, text=Success] at (12.08,5.23) {+21.6};
      \node[anchor=east, font=\tiny\bfseries, text=PaperMuted] at (8.16,4.44) {9B};
      \draw[-{Stealth[length=3.2pt]}, black!32, line width=0.65pt] (11.20,4.44) -- (12.93,4.44);
      \path[fill=black!13, draw=black!45, line width=0.5pt] (11.20,4.44) circle (0.088);
      \path[fill=MidAccentColor!85!white, draw=DarkAccentColor, line width=0.55pt] (13.08,4.44) circle (0.098);
      \node[anchor=east, font=\tiny\bfseries, text=black!55] at (11.04,4.44) {66.2};
      \node[anchor=west, font=\tiny\bfseries, text=DarkAccentColor] at (13.24,4.44) {89.8};
      \node[font=\tiny\bfseries, text=Success] at (12.14,4.63) {+23.6};
      \node[anchor=east, font=\tiny\bfseries, text=PaperMuted] at (8.16,3.50) {4B};
      \draw[-{Stealth[length=3.2pt]}, black!32, line width=0.65pt] (11.95,3.50) -- (12.76,3.50);
      \path[fill=black!13, draw=black!45, line width=0.5pt] (11.95,3.50) circle (0.088);
      \path[fill=MidAccentColor!85!white, draw=DarkAccentColor, line width=0.55pt] (12.91,3.50) circle (0.098);
      \node[anchor=east, font=\tiny\bfseries, text=black!55] at (11.79,3.50) {75.6};
      \node[anchor=west, font=\tiny\bfseries, text=DarkAccentColor] at (13.07,3.50) {87.6};
      \node[font=\tiny\bfseries, text=Success] at (12.43,3.69) {+12.0};
      \node[anchor=east, font=\tiny\bfseries, text=PaperMuted] at (8.16,2.90) {9B};
      \draw[-{Stealth[length=3.2pt]}, black!32, line width=0.65pt] (12.25,2.90) -- (12.92,2.90);
      \path[fill=black!13, draw=black!45, line width=0.5pt] (12.25,2.90) circle (0.088);
      \path[fill=MidAccentColor!85!white, draw=DarkAccentColor, line width=0.55pt] (13.07,2.90) circle (0.098);
      \node[anchor=east, font=\tiny\bfseries, text=black!55] at (12.09,2.90) {79.4};
      \node[anchor=west, font=\tiny\bfseries, text=DarkAccentColor] at (13.23,2.90) {89.6};
      \node[font=\tiny\bfseries, text=Success] at (12.66,3.09) {+10.2};
      \node[anchor=east, font=\tiny\bfseries, text=PaperMuted] at (8.16,1.96) {4B};
      \draw[-{Stealth[length=3.2pt]}, black!32, line width=0.65pt] (9.04,1.96) -- (9.70,1.96);
      \path[fill=black!13, draw=black!45, line width=0.5pt] (9.04,1.96) circle (0.088);
      \path[fill=MidAccentColor!85!white, draw=DarkAccentColor, line width=0.55pt] (9.85,1.96) circle (0.098);
      \node[anchor=east, font=\tiny\bfseries, text=black!55] at (8.88,1.96) {39.2};
      \node[anchor=west, font=\tiny\bfseries, text=DarkAccentColor] at (10.01,1.96) {49.4};
      \node[font=\tiny\bfseries, text=Success] at (9.44,2.15) {+10.2};
      \node[anchor=east, font=\tiny\bfseries, text=PaperMuted] at (8.16,1.36) {9B};
      \draw[-{Stealth[length=3.2pt]}, black!32, line width=0.65pt] (10.14,1.36) -- (11.73,1.36);
      \path[fill=black!13, draw=black!45, line width=0.5pt] (10.14,1.36) circle (0.088);
      \path[fill=MidAccentColor!85!white, draw=DarkAccentColor, line width=0.55pt] (11.88,1.36) circle (0.098);
      \node[anchor=east, font=\tiny\bfseries, text=black!55] at (9.98,1.36) {53.0};
      \node[anchor=west, font=\tiny\bfseries, text=DarkAccentColor] at (12.04,1.36) {74.8};
      \node[font=\tiny\bfseries, text=Success] at (11.01,1.55) {+21.8};
    \end{tikzpicture}}\end{minipage}
	\caption{\textbf{(Left)} A case of replacing the question image with its symbolic
	view, under the same model, question, and decoding procedure. \textbf{(Right)} The
	\emph{Symbolic Visual Gap}: symbolic views outperform visual inputs on every
	model and dataset pair from 4B to 9B.}
  \label{fig:symbolic-visual-gap}
\end{figure}
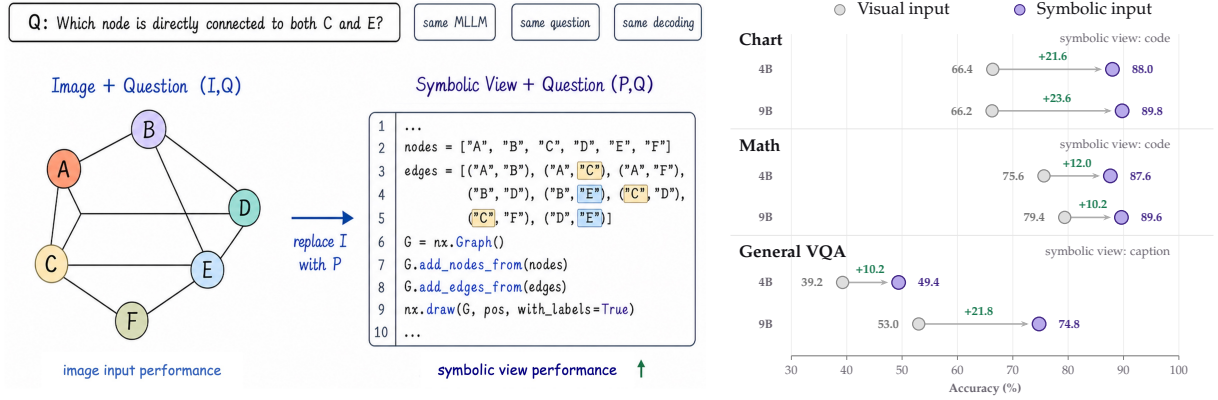

However, in this paper, we observe a counterintuitive phenomenon contradicting the seamless perception-reasoning transition.
As in \cref{fig:symbolic-visual-gap}, we use the generating code (or descriptive caption, which we term the \emph{symbolic view}) to replace the question image as the model input, with the model, question, and the decoding fixed, and measure the reasoning performance.
Since the input image is simple enough for perception, the image is ideally itself easier to understand than the generating code, making the replacement redundant.
Surprisingly, such a seemingly redundant operation significantly improves performance by 10.2\% to 23.6\% across model scales and datasets (\cref{fig:symbolic-visual-gap}, right).
This phenomenon makes us question the widely assumed seamless transition from perception to reasoning in visual reasoning: \textit{do we need another symbolic-view step during visual reasoning?}

To answer this question, we refer to this discrepancy between image-conditioned reasoning and symbol-conditioned reasoning as the \textbf{\emph{Symbolic Visual Gap}} and take a step closer to it.
We first analyze the visual information expressed in the generated reasoning traces for image-input models.
We find that the correct visual evidence can already appear in the trace, indicating that the model can already correctly perceive the visual information.
However, we still find that the model tends to adopt distracting evidence (visual evidence leading to incorrect reasoning and answers) for its subsequent reasoning, showing a disadvantage in visual evidence selection.
To verify it, we then quantitatively measure the answer-to-evidence attention between the image input and the symbolic-view input.
We find that answer-to-evidence attention favors distracting evidence given the image input, and correct evidence given the symbolic-view input, verifying the difficulty in selecting correct evidence for the image-input model.
In other words, despite good capabilities from current works in perception and reasoning themselves, \textbf{another bottleneck exists between perception and reasoning in selecting perceived visual information as appropriate evidence for subsequent reasoning}.

To handle this bottleneck, based on the above analysis, as the symbolic view elicits the model to assign answer-to-evidence attention to correct visual evidence, this view naturally offers a direction for guiding the image-input model for evidence selection without manually labeling correct evidence.
Moreover, since the symbolic view (generating code or descriptive caption) is easily obtained automatically, it is scalable for different data sizes.
Therefore, we propose to take the symbolic view as auxiliary information and introduce \mmopd, a multimodal on-policy self-distillation framework for correcting visual-evidence selection.
\mmopd takes the symbolic-view-input model as the teacher and uses the image-input model as the student, and then provides token-level supervision along trajectories generated by the student to guide the student to learn the evidence selection from the self-teacher.
During inference, since the symbolic view is not available, only the image-input model, with the symbolic visual gap largely reduced, will be used for visual reasoning.
\mmopd improves a broad range of multimodal abilities rather than a single targeted capability. Experiments across diverse benchmarks show consistent gains in visual perception, chart and document understanding, mathematical reasoning, and general multimodal reasoning.
Our contributions are as follows:

\begin{itemize}
  \item We identify the \emph{Symbolic Visual Gap}: with the model, question, and decoding fixed, replacing a visual input with its symbolic view improves accuracy by $10.2$ to $23.6$ percentage points across model scales and visual domains.

  \item We identify visual evidence selection as another critical bottleneck between perception and reasoning: correct visual information can appear in the generated trace without guiding the derivation, while image and symbolic conditioning exhibit opposite answer-to-evidence attention patterns.

  \item Based on this insight, we introduce \mmopd, a multimodal on-policy self-distillation framework, to transfer guidance from symbolic-conditioned behavior to the image-conditioned policy through residual token-level targets to help the model draw on correct visual evidence.

  \item Comprehensive experiments validate the effectiveness of \mmopd, demonstrating consistent improvements across model scales and diverse multimodal reasoning tasks.
\end{itemize}

\section{Delve into the Symbolic Visual Gap}
\label{sec:analysis}

The counterintuitive Symbolic Visual Gap inspires us to examine the assumed
seamless transition from perception to reasoning more closely.
In an ideal visual reasoning process, perceived information should become
appropriate evidence for subsequent reasoning.
We therefore examine (i) whether task-relevant visual information is correctly
perceived and represented as language-accessible evidence, and (ii) whether
subsequent reasoning draws on the correct evidence.

\subsection{From Visual Perception to Evidence Selection}
\label{sec:analysis-prelim}

When input an image into the MLLM (i.e., the image-input model), in the model-generated reasoning, some tokens verbalize information obtained
from the input image that may be used to answer the question. We refer to these as
\emph{visual evidence tokens}. For a given question, \emph{correct evidence tokens}
express visual information that supports a correct answer, whereas
\emph{distracting evidence tokens} express competing information that can
mislead subsequent reasoning or answer generation.
To distinguish visual perception from evidence use, we examine whether correct visual information already appears in the generated trace. Its explicit presence indicates that this information is available in a language-accessible form. We then inspect which evidence is used in subsequent reasoning and compare answer-to-evidence attention for correct and distracting evidence, thereby characterizing how strongly answer generation is associated with
each type of candidate evidence.

\begin{figure}[t]
  \centering
  \includegraphics[width=\linewidth]{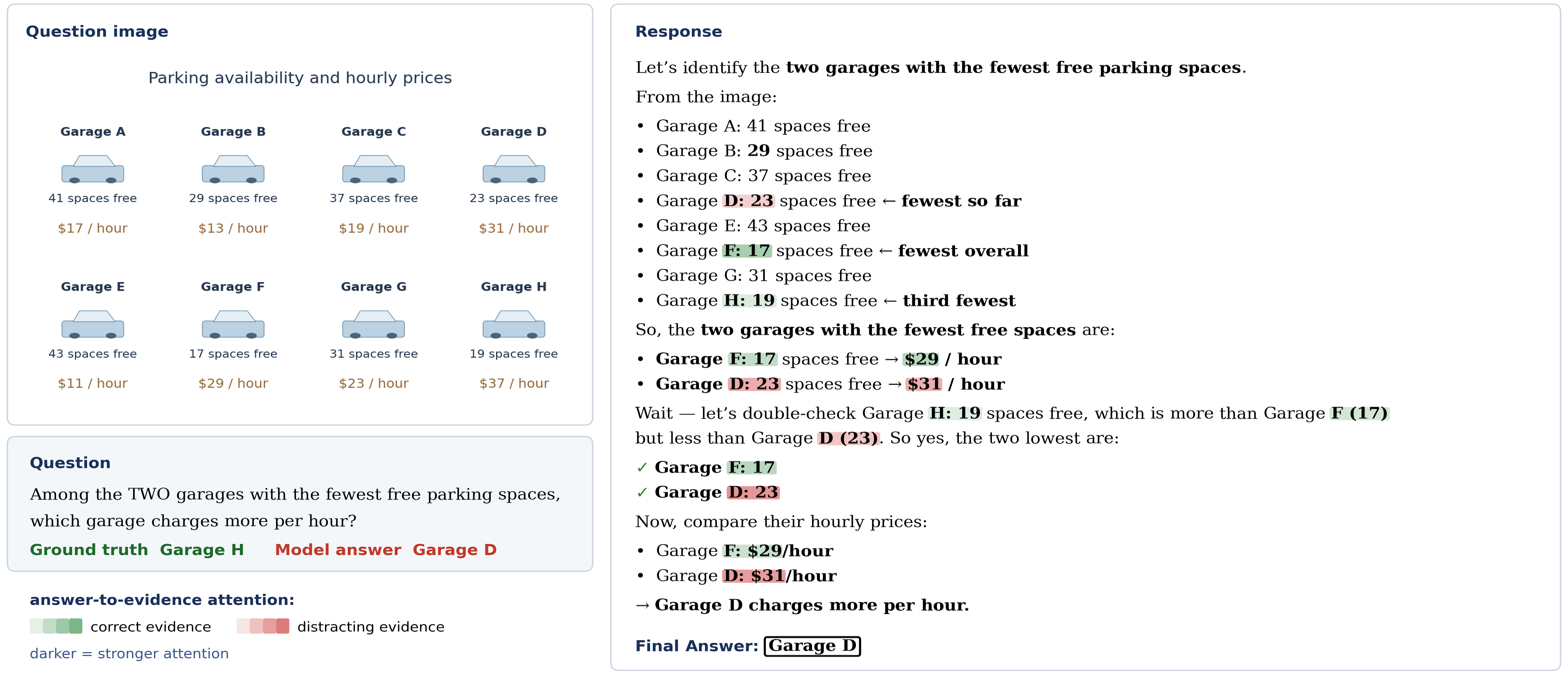}
  \caption{A case study. Task-relevant visual information is correctly perceived
    and verbalized, yet distracting evidence receives more attention and is selected
    as the premise for subsequent reasoning.}
  \label{fig:evidence-case}
\end{figure}

\Cref{fig:evidence-case} illustrates that correctly perceived visual
information may not be selected as appropriate evidence for subsequent reasoning.
The relevant values and their ordering are explicitly stated in the trace,
yet the model still selects distracting evidence as a premise. The subsequent
comparison is valid given that premise.
Distracting evidence also receives more attention,
including the evidence used in the derivation.
Thus, having relevant visual information available in the trace does not
ensure that it is used appropriately to answer the question.
Additional cases are provided in Appendix~\ref{sec:appendix-cases}.

\begin{takeaway}[Finding 1]
Correctly perceiving and verbalizing task-relevant visual information does not ensure correct evidence use. The model may select distracting evidence for subsequent reasoning.
\end{takeaway}

\subsection{Answer-to-Evidence Attention under Image and Symbolic Conditions}
\label{sec:analysis-quantitative}

We next quantitatively examine whether the attention preference observed in the case study
extends beyond this example and how it differs between image and symbolic
conditioning.
Specifically, we sample $200$ examples from each of three task families
(general VQA, chart reasoning, and mathematical reasoning), yielding $600$
examples in total.
Using Qwen3.5-4B, we measure at each full-attention layer the head-averaged
attention from the answer token to correct and distracting evidence tokens
under image and symbolic conditioning, and report both layer-wise and
overall results.

\begin{figure}[t]
  \centering
  \includegraphics[width=\linewidth]{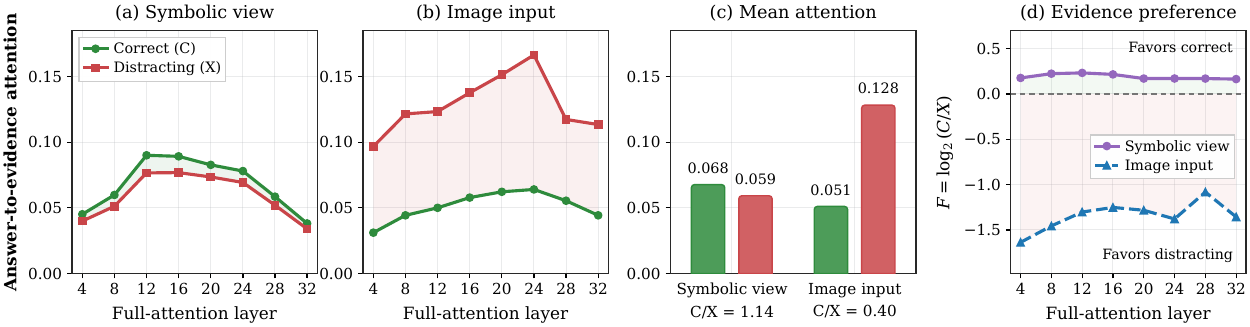}
  \caption{\textbf{Answer-to-evidence attention under image and symbolic conditioning.} (a,b) Layer-wise results at the observed full-attention layers. (c) Overall mean attention to correct and distracting evidence, together with their ratio. (d) The signed score $F=\log_2(C/X)$; positive and negative values indicate stronger association with correct and distracting evidence, respectively.}
  \label{fig:evidence-attention}
\end{figure}

\textbf{i) The two conditions exhibit opposite evidence orderings across depth.}
As in \cref{fig:evidence-attention}(a,b), distracting evidence receives
more attention than correct evidence at every observed layer under image
conditioning.
Symbolic conditioning consistently exhibits the opposite ordering.
This reversal persists across all observed layers and is therefore not
driven by an isolated layer.

\textbf{ii) The overall means and evidence ratios preserve the same trend.}
The overall statistics in \cref{fig:evidence-attention}(c) agree with the
layer-wise results.
Let $C$ and $X$ denote the mean attention to correct and distracting evidence,
respectively. We define their correct-to-distracting ratio as
\begin{equation}
  R=\frac{C}{X},
  \label{eq:evidence-ratio}
\end{equation}
where $R>1.0$ and $R<1.0$ indicate stronger association with correct and distracting
evidence, respectively.
The ratio is above $1.0$ under symbolic conditioning and below $1.0$ under image
conditioning.

\textbf{iii) A signed score summarizes the reversal.}
For a signed summary of the same relative attention, we define
\begin{equation}
  F=\log_2 R=\log_2(C/X).
  \label{eq:evidence-favor}
\end{equation}
Positive values indicate a stronger association with correct evidence,
whereas negative values indicate a stronger association with distracting
evidence.
As shown in \cref{fig:evidence-attention}(d), $F$ is positive at every observed
layer under symbolic conditioning and negative at every observed layer under
image conditioning. The overall scores retain the same signs.

Together with the case study, the consistent attention preference for
distracting evidence under image conditioning points to a difficulty in
using perceived information as evidence for reasoning.

\begin{takeaway}[Finding 2]
Visual evidence selection is one more bottleneck between perception and reasoning. Beyond perception and reasoning, effective multimodal capability also depends on whether perceived information is selected as appropriate evidence for subsequent reasoning.
\end{takeaway}

\subsection{Summary and Discussion}
\label{sec:analysis-insight}

The above analyses challenge the assumed seamless transition from perception
to reasoning. Relevant visual information can already appear in the generated
trace, yet the model may still select distracting evidence for subsequent
reasoning. In other words, perceiving the required information does not ensure
that it becomes appropriate evidence for answering the question.

To address this bottleneck, the comparison with symbolic views offers a
natural direction. With the model and question fixed, symbolic conditioning
elicits stronger performance and an attention preference for correct evidence.
This contrast suggests that the behavior elicited by symbolic views can guide
evidence selection under image conditioning. Moreover, symbolic views,
instantiated as generating code or descriptive captions, can be obtained
automatically at scale, providing a scalable source of guidance without
manually labeling correct evidence.

\begin{takeaway}[Insight]
Symbolic Visual Gap reveals a missing link between perception and reasoning: visual information may be perceived without being correctly selected. Symbolic views elicit better evidence selection that can guide image-conditioned reasoning without manual evidence labeling.
\end{takeaway}

\section{Method}
\label{sec:method2}

\begin{figure}[!t]
	\centering
	\includegraphics[width=\linewidth]{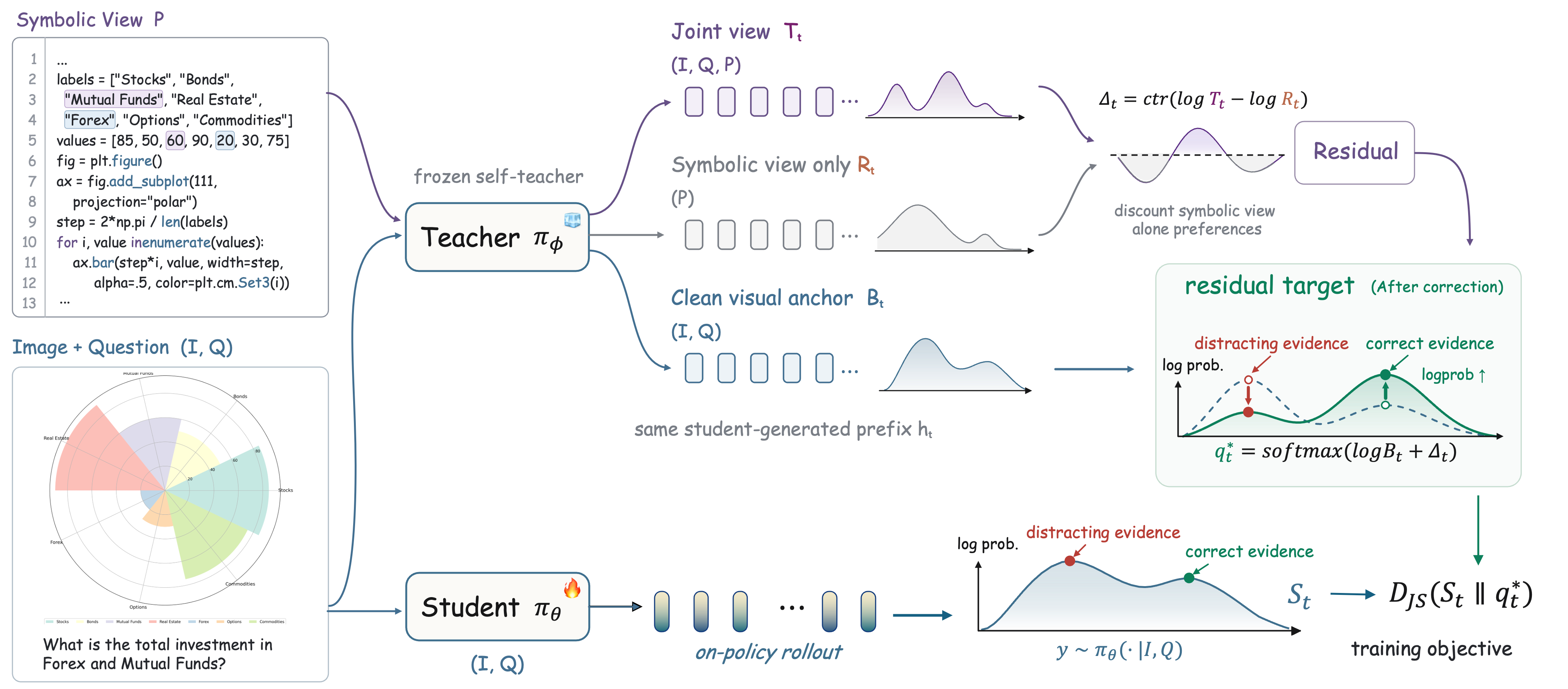}
	\caption{Overview of \mmopd, a multimodal OPSD
	framework for symbolic-to-visual correction. Symbolic views provide
	privileged supervision, while residual targets discount token preferences
	induced by these views alone and guide image-conditioned reasoning toward
	correct visual evidence.}
	\label{Fig.intro} \end{figure}

The above analysis reveals visual evidence selection as one more bottleneck between perception and reasoning. It also shows that symbolic views steer answer-to-evidence attention toward correct visual evidence. Since these views can be obtained automatically at scale, this finding motivates us to introduce \mmopd, a multimodal on-policy self-distillation framework for symbolic-to-visual correction. A frozen self-teacher uses additional symbolic information to supervise the student along student-generated trajectories. Residual token-level targets reduce the influence of token preferences induced by the symbolic view alone and refine image-conditioned predictions.
The overview framework is shown in~\Cref{Fig.intro}, with training pseudocode in~\cref{alg:mmopd2}

\subsection{Symbolic-to-Visual Correction}
\label{sec:method2-alignment}

Given the curated dataset $\mathcal{D}=\{(I_i,Q_i,P_i)\}_{i=1}^{N}$, where $P_i$ is privileged symbolic information, we initialize a trainable student
$\pi_\theta$ and a frozen self-teacher $\pi_\phi$ from the same pretrained
MLLM checkpoint. For each
$(I,Q,P)\sim\mathcal D$, the image-conditioned student generates an on-policy
response:
\begin{equation}
	y=(y_1,\ldots,y_{|y|})
	\sim \pi_\theta(\cdot\mid I,Q),
	\qquad
	h_t=y_{<t},
	\label{eq:student-rollout}
\end{equation}
where $h_t$ is the student-generated prefix at position $t$. Let $\mathcal V$
denote the model vocabulary and $v\in\mathcal V$ a candidate next token. We
evaluate four distributions on the same prefix:
\begin{equation}
	\begin{aligned}
		S_t(v) &= \pi_\theta(v\mid I,Q,h_t),
		&
		T_t(v) &= \pi_\phi(v\mid I,Q,P,h_t),\\
		B_t(v) &= \pi_\phi(v\mid I,Q,h_t),
		&
		R_t(v) &= \pi_\phi(v\mid P,h_t).
	\end{aligned}
	\label{eq:four-model-views}
\end{equation}
Here, $S_t$ is the current student prediction. The other three distributions
are obtained from the single frozen self-teacher under different conditioning
views: $T_t$ combines the image, question, and privileged symbolic information;
$B_t$ is the clean image-conditioned prediction; and $R_t$ estimates the preference induced by $P$ without direct access
to the image or question. Their differences reflect the conditioning view rather than different generated trajectories.

\paragraph{Residual target construction}
Directly distilling $T_t$ would conflate the task-conditioned benefit of $P$
with token preferences that $P$ can induce on its own. \mmopd therefore uses
$R_t$ as a privilege-only reference and contrasts it with $T_t$. For the same
token, the log-ratio $\log T_t(v)-\log R_t(v)$ discounts the preference
explained by $P$ alone, leaving a correction associated with interpreting $P$
in the full image--question context.

For tractable token-level distillation, the privileged teacher defines a
shared candidate set $\mathcal K_t=\operatorname{Top}_{K}(T_t)$ containing its
$K$ most probable tokens; probabilities outside this set are later aggregated
into a tail bucket. We remove the vocabulary-wide offset from the log-ratio:
\begin{equation}
	\widetilde d_t(v)
	=
	\log T_t(v)-\log R_t(v)
	-
	\mathbb E_{u\in\mathcal V}
	\left[
		\log T_t(u)-\log R_t(u)
	\right].
	\label{eq:privileged-residual}
\end{equation}
The expectation is uniform over $\mathcal V$. Subtracting it removes a
token-independent shift, which carries no information about the relative
preference among candidate tokens. Thus, $\widetilde d_t(v)$ measures how the
full task context changes the relative preference for $v$ beyond what is
induced by $P$ alone.

Rather than replacing the image-conditioned prediction with the privileged
teacher, we use this residual to correct the clean anchor $B_t$:
\begin{equation}
	\widehat q_t
	=
	\operatorname{softmax}_{\mathcal K_t}
	\left(
		\log B_t
		+
		\gamma\tanh\left(\frac{\widetilde d_t}{\gamma}\right)
	\right).
	\label{eq:image-anchored-head}
\end{equation}
Adding the residual in log-probability space reweights the token preferences of
$B_t$, so privileged information acts as a correction to the prediction made
from the actual inference input. The saturation function remains approximately
linear for moderate residuals while suppressing extreme log-ratios, with
$\gamma>0$ controlling its scale.

Finally, let $M_{B,t}=\sum_{v\in\mathcal K_t}B_t(v)$ be the probability mass
that the clean anchor assigns to the selected support. We preserve this mass
and distribute it according to $\widehat q_t$:
\begin{equation}
	q_t^\star(v)
	=
	M_{B,t}\widehat q_t(v),
	\qquad v\in\mathcal K_t.
	\label{eq:method2-image-anchored-target}
\end{equation}
Writing $\mathrm{tail}$ for all tokens outside $\mathcal K_t$, we retain the
remaining anchor mass:
\begin{equation}
	q_t^\star(\mathrm{tail})
	=
	1-M_{B,t}
	=
	B_t(\mathrm{tail}).
	\label{eq:method2-image-anchored-tail}
\end{equation}
Hence, the residual changes only the relative probabilities within the
teacher-selected support; the total Top-$K$ and tail masses remain grounded in
the clean image-conditioned prediction.

\subsection{Training Objective}
\label{sec:training-objective}

The target $q_t^\star$ is defined on the shared support
$\mathcal K_t\cup\{\mathrm{tail}\}$. To evaluate the student on the same
coordinates, we retain its probabilities on $\mathcal K_t$ and aggregate all
remaining probability mass into a single tail bucket:
\begin{equation}
	\widetilde S_t(v)=S_t(v)
	\quad (v\in\mathcal K_t),
	\qquad
	\widetilde S_t(\mathrm{tail})
	=
	1-\sum_{u\in\mathcal K_t}S_t(u).
	\label{eq:method2-student-tail}
\end{equation}
Both $\widetilde S_t$ and $q_t^\star$ are therefore normalized distributions
on the same support. Let $M_t=\frac{1}{2}(\widetilde S_t+q_t^\star)$ denote
their equal-weight mixture. At each response position, we minimize the
symmetric Jensen--Shannon divergence
\begin{equation}
	D_{\mathrm{JS}}
	\left(
	\widetilde S_t\,\middle\|\,q_t^\star
	\right)
	=
	\frac{1}{2}
	D_{\mathrm{KL}}
	\left(
	\widetilde S_t\,\middle\|\,M_t
	\right)
	+
	\frac{1}{2}
	D_{\mathrm{KL}}
	\left(
	q_t^\star\,\middle\|\,M_t
	\right).
	\label{eq:js-distillation}
\end{equation}
The overall training objective averages this dense token-level signal over
both the curated data and on-policy rollouts:
\begin{equation}
	\mathcal L_{\mathrm{MM\text{-}OPD}}(\theta)
	=
	\mathbb E_{\substack{
			(I,Q,P)\sim\mathcal D\\
			y\sim\pi_\theta(\cdot\mid I,Q)
	}}
	\left[
	\frac{1}{|y|}
	\sum_{t=1}^{|y|}
	D_{\mathrm{JS}}
	\left(
	\widetilde S_t\,\middle\|\,q_t^\star
	\right)
	\right].
	\label{eq:method2-objective}
\end{equation}
The teacher-side distributions $T_t$, $B_t$, and $R_t$ are detached, and
$q_t^\star$ is treated as a fixed target; gradients flow only through the
image-conditioned student. Since each prefix is generated by the student
itself, the objective supplies token-level supervision on the state
distribution encountered at inference time.

\begin{algorithm}[t]
	\caption{\mmopd Training}
	\label{alg:mmopd2}
	\begin{algorithmic}[1]
		\Require Student $\pi_\theta$, frozen self-teacher
		$\pi_\phi$, training sample $(I,Q,P)$, rollout number $n$, residual
		saturation scale $\gamma$
		\State Sample $n$ independent on-policy responses
		$\mathcal Y=\{y^{(j)}\}_{j=1}^{n}$ from $\pi_\theta(\cdot\mid I,Q)$
		\Comment{\cref{eq:student-rollout}}
		\For{each $y\in\mathcal Y$ and position $t$ with prefix $h_t=y_{<t}$}
			\State Evaluate $S_t$, $T_t$, $B_t$, and $R_t$ on $h_t$
			\Comment{\cref{eq:four-model-views}}
			\State Select
			$\mathcal K_t\gets\operatorname{Top}_{K}(T_t)$
			\State Extract the centered privileged residual $\widetilde d_t$
			\Comment{\cref{eq:privileged-residual}}
			\State Construct the residual-corrected distribution $\widehat q_t$
			\Comment{\cref{eq:image-anchored-head}}
			\State Preserve the anchor's head and tail masses to obtain
			$q_t^\star$
			\Comment{\cref{eq:method2-image-anchored-target,eq:method2-image-anchored-tail}}
			\State Compress $S_t$ onto
			$\mathcal K_t\cup\{\mathrm{tail}\}$ as $\widetilde S_t$
			\Comment{\cref{eq:method2-student-tail}}
		\EndFor
		\State Update $\theta$ by minimizing
		$\mathcal L_{\mathrm{MM\text{-}OPD}}(\theta)$
		\Comment{\cref{eq:method2-objective}}
	\end{algorithmic}
\end{algorithm}

\paragraph{Training and inference}
Training draws $n$ independent on-policy student rollouts per prompt and performs three gradient-free teacher
evaluations of each trajectory under the joint privileged, clean visual, and
privilege-only views. At inference, the symbolic information and all
teacher-side evaluations are removed; the trained student decodes normally
from $(I,Q)$ without additional inputs or computation.

\subsection{Training Dataset Curation}
\label{sec:training-data-curation}

To turn the Symbolic Visual Gap into a training signal, we construct
$\mathcal{D}=\{(I_i,Q_i,P_i)\}_{i=1}^{N}$, where $P_i$ is privileged symbolic information, instantiated as a
caption for a natural image or executable code for a structured visual. We use
the following three-stage curation pipeline.

\textbf{Cross-View Validation.}
We sample paired examples from V-Interaction~\citep{qiao2025v}, Mulberry~\citep{yao2026mulberry},
and ChartX~\citep{xia2025chartx}, covering general VQA, mathematical reasoning,
and chart understanding. The corresponding captions and code are taken
directly from the source datasets. For natural images, we discard caption--image pairs
whose contents are semantically inconsistent. For structured visuals, we
execute the code and discard any sample whose code fails to execute,
including those with syntax errors or missing dependencies.

\textbf{Symbolic-View Normalization.}
The dataset-provided captions and code follow heterogeneous formats. We use Qwen3.5-397B to reorganize and normalize these existing symbolic views while preserving their original visual content; no new caption or code content is generated. Existing captions are reorganized into a consistent descriptive format, with boilerplate and image-irrelevant content removed. For code, we remove environment-specific elements, including hard-coded file paths and output operations such as \texttt{savefig}, while retaining the statements that construct the visual. This step reduces formatting variation and removes environment-dependent code across data sources.

\textbf{Symbolic-View Utility Filtering.}
A valid image--symbolic pair does not necessarily improve the model's ability
to answer its question. We estimate
$\operatorname{Pass@8}$ with the original visual input $(I,Q)$ and with the
additional symbolic view $(I,P,Q)$. We retain only examples satisfying
$\operatorname{Pass@8}(I,Q)\leq\operatorname{Pass@8}(I,P,Q)$, excluding cases
in which access to $P$ lowers $\operatorname{Pass@8}$. The resulting dataset
contains 32K examples spanning general VQA, mathematical diagrams, and charts.

\section{Experiments}
\label{sec:experiments}

\subsection{Experimental Setup}
\label{sec:exp-setup}

\textbf{Models and training data.}
We apply \mmopd to Qwen3.5-4B and Qwen3.5-9B~\citep{qwen3.5}, both trained on
\textsc{MMOPD-32K}, which pairs each image-question example with a symbolic
view: a descriptive caption for a natural image or executable code for a
mathematical diagram or chart. The symbolic view is used only during training
and in the controlled gap analysis; all standard benchmark evaluations use
only the original image and question. The construction and filtering of the
training set are described in \cref{sec:training-data-curation}.

\textbf{Implementation details.}
Unless stated otherwise, \mmopd uses a frozen self-teacher initialized from
the student checkpoint and the Jensen-Shannon divergence (JSD) as the
distillation objective. Token-level distillation is computed over the top-$K$
support with $K=100$, and the residual saturation scale is set to
$\gamma=10$. Each prompt produces $n=2$ on-policy rollouts with a maximum
response length of $8192$ tokens. We train for one epoch with a global batch
size of $256$ and a learning rate of $1\mathrm{e}{-}6$ on $8\times$
$141$\,GB H20 GPUs. Controlled baselines reuse the same backbone, training
data, and prompts; we match the number of updates and the generated-token
budget where applicable while keeping method-specific optimization settings.
At inference, \mmopd requires neither the symbolic view nor the self-teacher.

\textbf{Benchmarks and metrics.}
We evaluate four groups of capabilities. \emph{General VQA} includes
MMStar~\citep{chen2024we}, RealWorldQA~\citep{grok15v}, MMBench
v1.1~\citep{liu2024mmbench}, and SimpleVQA~\citep{cheng2025simplevqa}.
\emph{Multimodal reasoning} includes MathVista-mini~\citep{lu2024mathvista},
MathVerse-mini~\citep{zhang2024mathverse}, DynaMath~\citep{zou2025dynamath},
We-Math~\citep{qiao2025we}, LogicVista~\citep{xiao2024logicvista},
MMMU~\citep{yue2024mmmu}, and MMMU-Pro~\citep{yue2025mmmu}.
\emph{Fine-grained perception} includes V* Bench~\citep{wu2024v} and the 4K
and 8K splits of HR-Bench~\citep{wang2025divide}. \emph{Document and chart
understanding} includes ChartQA~\citep{masry2022chartqa}, the descriptive (DQ)
and reasoning (RQ) tracks of CharXiv~\citep{wang2024charxiv},
DocVQA~\citep{mathew2021docvqa}, OCRBench~\citep{liu2024ocrbench}, and the
English-task average of OCRBench v2~\citep{fu2026ocrbench}.

\textbf{Compared methods.}
All controlled methods start from the corresponding Qwen3.5 checkpoint and share
the same training dataset MMOPD-32K. \textbf{Base} denotes the unchanged
model. \textbf{SFT} performs teacher-forced fine-tuning on reference
responses~\citep{ouyang2022training}. \textbf{GRPO}~\citep{shao2024deepseekmath},
\textbf{GSPO}~\citep{zheng2025gspo}, and \textbf{DAPO}~\citep{yu2026dapo} are
RLVR baselines trained with the same judge Qwen3-32B. \textbf{Vanilla OPSD}~\citep{zhao2026self} transplants on-policy self-distillation to the
multimodal setting in its original form: the frozen self-teacher is conditioned
on the reference answer as privileged information, and the student model distills the resulting teacher distribution over its own rollouts. It therefore shares the on-policy sampling procedure, the frozen self-teacher, and the top-$K$ support with \mmopd, and differs both in what privileges the teacher and in how the distillation target is formed. The three families are separated by the supervision they consume: SFT and RLVR take the reference answer as a training target or a verifier label, vanilla OPSD conditions its teacher on the reference answer, and \mmopd uses only the symbolic view $P$ and never accesses the reference answer.

\textbf{Evaluation protocol.}
For a fair comparison, all checkpoints are evaluated with
VLMEvalKit~\citep{duan2024vlmevalkit} using the same prompts, image
preprocessing, answer extraction, and benchmark-specific metrics. For
benchmarks requiring an LLM judge, we use Qwen3-32B with identical judging
prompts across all methods. This unified protocol controls evaluation-related
variation across methods.

\subsection{Main Results}
\label{sec:exp-results}

We first compare \mmopd with controlled post-training baselines across all
benchmarks.

{\setlength{\intextsep}{16pt plus 2pt minus 2pt}
\begin{table}[t]
	\renewcommand{\best}[1]{\textbf{#1}}
  \centering
  \captionsetup{skip=8pt}
  \caption{\textbf{Main results across diverse multimodal benchmarks.} We compare
  different post-training methods on the Qwen3.5-4B and Qwen3.5-9B backbones.
  Benchmarks are grouped into general VQA, reasoning, perception, and document
  understanding. Results are obtained with VLMEvalKit using each benchmark's
  evaluation metric. The best result in is shown in \best{bold} and the second best is \second{underlined}.}
  \label{tab:main-results}
  \scriptsize
  \setlength{\tabcolsep}{2.5pt}
  \setlength{\aboverulesep}{0pt}
  \setlength{\belowrulesep}{0pt}
  \renewcommand{\arraystretch}{1.55}
  \resizebox{\textwidth}{!}{\begin{tabular}{@{}c|c|*{7}{c}|*{7}{c}@{}}
    \toprule
    \multirow{2}{*}{\textbf{Category}} &
    \multirow{2}{*}{\textbf{Benchmark}} &
    \multicolumn{7}{c|}{\textbf{Qwen3.5-4B}} &
    \multicolumn{7}{c}{\textbf{Qwen3.5-9B}} \\
    \cmidrule(lr){3-9}\cmidrule(lr){10-16}
    & & \textbf{Baseline} & \textbf{ SFT } & \textbf{GRPO} & \textbf{GSPO} &
    \textbf{DAPO} & \textbf{OPSD} & {\fontsize{8.5}{9}\selectfont\mmopd} &
    \textbf{Baseline} & \textbf{ SFT } & \textbf{GRPO} & \textbf{GSPO} &
    \textbf{DAPO} & \textbf{OPSD} & {\fontsize{8.5}{9}\selectfont\mmopd} \\
    \midrule
    \multirow{4}{*}{\shortstack{General\\VQA}}
      & MMStar           & 72.9           & 70.1           & \second{73.8}  & 72.1           & 68.7           & 72.8           & \best{74.6}    & 74.9           & 70.5           & 73.4           & 74.5           & 73.0           & \second{75.4}  & \best{75.9} \\
      & RealWorldQA      & 75.7           & 75.4           & \best{77.8}    & 76.5           & 74.2           & 76.1           & \second{77.4}  & \second{78.0}  & \best{78.3}    & 76.1           & 76.2           & 76.1           & 77.9           & 77.5 \\
      & MMBench v1.1     & 85.7           & 83.5           & \second{86.5}  & 86.3           & 84.8           & 85.2           & \best{86.6}    & \second{87.2}  & 84.4           & 86.1           & 85.7           & 86.0           & \second{87.2}  & \best{87.7} \\
      & SimpleVQA        & 50.7           & 49.9           & 51.2           & 47.5           & \best{52.2}    & 48.4           & \second{51.7}  & 51.6           & 49.2           & 52.4           & \best{53.3}    & 52.9           & \second{53.2}  & \second{53.2} \\
    \midrule
    \multirow{7}{*}{\shortstack{Reasoning\\Tasks}}
      & MathVista$_{mini}$ & 81.4           & 75.9           & \second{81.8}  & 80.5           & 75.7           & \second{81.8}  & \best{82.2}    & 82.9           & 77.9           & 83.2           & 82.9           & \second{83.4}  & 83.0           & \best{84.1} \\
      & MathVerse$_{mini}$ & 53.9           & 50.2           & 53.3           & 53.9           & \second{54.3}  & 53.4           & \best{54.5}    & 54.1           & 52.0           & 47.9           & 49.9           & \second{54.9}  & 52.9           & \best{57.9} \\
      & DynaMath         & \second{72.4}  & 70.0           & 71.9           & 71.5           & 70.3           & 70.9           & \best{74.0}    & 75.0           & 72.4           & 75.8           & 73.4           & 74.4           & \second{76.1}  & \best{77.7} \\
      & We-Math          & \second{71.1}  & 69.0           & 70.4           & 70.6           & 70.4           & \second{71.1}  & \best{74.4}    & 73.0           & 66.7           & \second{74.5}  & 73.5           & 70.2           & 74.2           & \best{76.5} \\
      & LogicVista       & 66.7           & 61.6           & 63.1           & 66.0           & 65.6           & \second{68.0}  & \best{69.4}    & 67.3           & 64.7           & 65.5           & 67.1           & 63.1           & \second{70.2}  & \best{73.4} \\
      & MMMU             & \second{72.6}  & 68.7           & 71.3           & 71.7           & 70.5           & 72.2           & \best{73.8}    & \second{74.7}  & 70.0           & 72.6           & \second{74.7}  & 71.1           & 74.1           & \best{75.8} \\
      & MMMU-Pro         & 57.6           & 53.6           & 56.3           & 57.3           & 55.9           & \best{59.2}    & \second{58.5}  & 62.7           & 55.1           & 59.5           & 59.9           & 58.0           & \second{64.1}  & \best{64.7} \\
    \midrule
    \multirow{3}{*}{\shortstack{Perception\\Tasks}}
      & V* Bench         & 81.7           & 76.4           & 83.8           & \second{84.3}  & 80.1           & \second{84.3}  & \best{88.5}    & 82.2           & 84.3           & 83.8           & 84.3           & 82.2           & \best{86.9}    & \second{84.8} \\
      & HR-Bench 4K      & 81.5           & 80.4           & \second{83.6}  & 81.6           & 78.0           & 80.3           & \best{85.0}    & \second{82.7}  & 80.9           & 82.3           & 82.4           & 80.0           & 81.3           & \best{84.0} \\
      & HR-Bench 8K      & 75.1           & 72.4           & 76.1           & \second{76.9}  & 73.3           & 76.0           & \best{80.5}    & 77.6           & 75.8           & 78.0           & 76.9           & 77.5           & \best{79.6}    & \second{78.4} \\
    \midrule
    \multirow{6}{*}{\shortstack{Document\\Understanding}}
      & ChartQA          & 84.2           & 80.4           & 83.9           & 76.4           & 80.2           & \second{84.4}  & \best{85.1}    & \second{85.4}  & 81.7           & 84.6           & 84.8           & 83.6           & 84.0           & \best{93.7} \\
      & CharXiv (DQ)     & 88.0           & 81.3           & \second{88.6}  & 85.6           & 87.6           & \best{88.9}    & 88.0           & \second{91.1}  & 85.9           & 89.8           & 88.0           & 90.0           & 90.7           & \best{91.8} \\
      & CharXiv (RQ)     & 65.9           & 62.8           & \second{67.6}  & 62.5           & 62.2           & 64.5           & \best{68.8}    & \second{69.0}  & 63.1           & 68.6           & 67.7           & 68.2           & 68.6           & \best{71.2} \\
      & DocVQA           & 95.1           & 94.2           & \second{95.3}  & 94.7           & 94.9           & 95.0           & \best{95.5}    & 94.9           & \best{95.4}    & 94.9           & 94.9           & \second{95.0}  & \second{95.0}  & \best{95.4} \\
      & OCRBench         & 87.1           & 81.3           & 87.4           & 86.5           & 86.4           & \second{87.6}  & \best{87.8}    & 85.9           & 84.0           & \best{87.3}    & 86.4           & 86.7           & 86.0           & \second{87.0} \\
      & OCRBench v2 (EN) & 53.9           & 48.9           & 53.8           & \second{55.0}  & 52.0           & 52.5           & \best{55.1}    & 56.3           & 47.5           & 56.5           & 56.4           & \second{56.8}  & 56.3           & \best{57.2} \\
    \midrule
    \textbf{Total} &\textbf{Average} & 73.7           & 70.3           & \second{73.9}  & 72.9           & 71.9           & 73.6           & \best{75.6}    & 75.3           & 72.0           & 74.6           & 74.6           & 74.2           & \second{75.8}  & \best{77.4} \\
    \bottomrule
\end{tabular}}
\end{table}
}

\textbf{Broad and consistent gains across capabilities.}
As shown in \cref{tab:main-results}, \mmopd achieves the highest average
score at both model scales, outperforming the Qwen3.5-4B and Qwen3.5-9B
baselines by $1.9$ and $2.1$ points, respectively. More importantly, it
improves performance in $38$ of the $40$ model--benchmark settings, covering
general VQA, multimodal reasoning, fine-grained perception, and document
understanding. These results demonstrate that the
benefits of \mmopd extend consistently across capabilities rather than
concentrating on a single task family.

\textbf{Stronger and more balanced than SFT and RLVR.}
With the same training data and backbones, SFT reduces the average score by
$3.4$ points at 4B and $3.3$ points at 9B. GRPO, GSPO, and DAPO obtain gains
on individual benchmarks but regress substantially on others; for instance,
GSPO decreases ChartQA by $7.8$ points at 4B, while GRPO decreases
MathVerse-mini by $6.2$ points at 9B. In contrast, \mmopd surpasses the best
average among the RLVR baselines by $1.7$ points at 4B and $2.8$ points at
9B. Thus, \mmopd avoids trading gains in one capability for degradation in another.

\textbf{Benefits over vanilla OPSD.}
Vanilla OPSD shares the on-policy sampling procedure and frozen self-teacher
with \mmopd, but differs in both the privileged supervision and the
distillation target: it directly distills an answer-conditioned teacher,
whereas \mmopd constructs a residual correction from the symbolic view.
\mmopd improves the average over vanilla OPSD by $2.0$ points at 4B and
$1.6$ points at 9B, including a $5.0$-point gain on MathVerse-mini and a
$9.7$-point gain on ChartQA at 9B. These results support the combined benefit
of symbolic supervision and residual target construction for multimodal
on-policy self-distillation.

\subsection{Ablation Study}
\label{sec:exp-ablation}

We conduct all ablations with Qwen3.5-4B on six benchmarks covering general VQA, multimodal reasoning, and fine-grained perception. Each experiment varies one factor while keeping the remaining settings at their defaults. We report individual benchmark scores and their unweighted average.

\begin{table}[H]
  \centering
	\caption{\textbf{Effect of the distillation objective on Qwen3.5-4B.}
	Forward and reverse KL denote
	$D_{\mathrm{KL}}(q_t^\star\|\widetilde S_t)$ and
	$D_{\mathrm{KL}}(\widetilde S_t\|q_t^\star)$, respectively.
	The residual target construction is unchanged.}
  \label{tab:divergence-ablation}
  \scriptsize
  \resizebox{\textwidth}{!}{\begin{tabular}{cccccccc}
    \toprule
    \textbf{Divergence} & \textbf{V* Bench} &
    \textbf{HR-Bench 4K} &
    \textbf{HR-Bench 8K} & \textbf{MMStar} &
    \textbf{MathVista}$_{\boldsymbol{mini}}$ & \textbf{LogicVista} &
    \textbf{Average} \\
    \midrule
    Forward KL   & 84.3          & 84.8          & 79.5          & 74.4          & 82.0          & 67.6          & 78.8 \\
    Reverse KL   & 86.4          & \textbf{85.3} & 79.6          & 74.5          & \textbf{82.7} & \textbf{70.0} & 79.8 \\
    \textbf{JSD} & \textbf{88.5} & 85.0          & \textbf{80.5} & \textbf{74.6} & 82.2          & 69.4          & \textbf{80.0} \\
    \bottomrule
  \end{tabular}}
\end{table}

\textbf{Distillation objective.}
As shown in \cref{tab:divergence-ablation}, we compare forward KL, reverse KL, and JSD as the token-wise divergence measure while keeping the residual target construction unchanged. All three objectives improve over the baseline on every benchmark, showing that the benefits of \mmopd are not specific to a single divergence function. JSD and reverse KL each lead on three benchmarks, indicating different preferences across tasks. JSD achieves the highest overall average and is therefore adopted as our default.

\begin{table}[H]
  \centering
  \caption{\textbf{Effect of the teacher update rule on Qwen3.5-4B.} Current uses the latest student weights, EMA tracks their exponential moving average, and Frozen retains the initial weights.}
  \label{tab:teacher-ablation}
  \scriptsize
  \resizebox{\textwidth}{!}{\begin{tabular}{cccccccc}
    \toprule
    \textbf{Teacher Update} & \textbf{V* Bench} &
    \textbf{HR-Bench 4K} &
    \textbf{HR-Bench 8K} & \textbf{MMStar} &
    \textbf{MathVista}$_{\boldsymbol{mini}}$ & \textbf{LogicVista} &
    \textbf{Average} \\
    \midrule
    EMA             & 81.1          & \textbf{85.8} & 79.8          & \textbf{75.0} & 81.6          & 67.1          & 78.4 \\
    \textbf{Frozen} & \textbf{88.5} & 85.0          & \textbf{80.5} & 74.6          & \textbf{82.2} & \textbf{69.4} & \textbf{80.0} \\
    \bottomrule
  \end{tabular}}
\end{table}

\textbf{Teacher update rule.}
We examine how updating the self-teacher affects distillation (\cref{tab:teacher-ablation}). Both EMA and frozen teachers improve over the baseline on average. EMA performs better on HR-Bench 4K and MMStar, whereas the frozen teacher leads on the remaining four benchmarks and achieves a $1.6$-point higher average. We therefore keep the teacher parameters frozen at initialization during training.

\begin{table}[H]
  \centering
  \caption{\textbf{Effect of the maximum on-policy response length.} We vary
  the response length on Qwen3.5-4B while fixing the other settings as default.}
  \label{tab:length-ablation}
  \scriptsize
  \resizebox{\textwidth}{!}{\begin{tabular}{cccccccc}
    \toprule
    \textbf{Response Length} & \textbf{V* Bench} &
    \textbf{HR-Bench 4K} &
    \textbf{HR-Bench 8K} & \textbf{MMStar} &
    \textbf{MathVista}$_{\boldsymbol{mini}}$ & \textbf{LogicVista} &
    \textbf{Average} \\
    \midrule
    1024          & 84.3          & 83.5          & 78.6          & 75.0          & 81.4          & 66.2          & 78.2 \\
    2048          & 84.8          & 84.1          & 79.1          & \textbf{75.4} & 81.7          & 66.4          & 78.6 \\
    4096          & 86.4          & 84.2          & 79.9          & 73.9          & 82.1          & \textbf{69.6} & 79.4 \\
    \textbf{8192} & \textbf{88.5} & 85.0          & \textbf{80.5} & 74.6          & 82.2          & 69.4          & \textbf{80.0} \\
    16384         & 86.9          & \textbf{85.2} & \textbf{80.5} & 74.3          & \textbf{82.7} & 68.1          & 79.6 \\
    \bottomrule
  \end{tabular}}
\end{table}

\textbf{Maximum response length.}
As shown in \cref{tab:length-ablation}, we vary the maximum length of on-policy training responses from $1024$ to $16384$ tokens. Different benchmarks favor different length limits, and increasing the limit does not consistently improve performance on every task. Nevertheless, all tested limits outperform the baseline on average, demonstrating robust gains across the evaluated range. Average performance peaks at $8192$ and declines slightly at $16384$; we therefore adopt $8192$ as the default.

\begin{table}[H]
  \centering
\caption{\textbf{Effect of the number of on-policy rollouts.}
	We vary the number of rollouts per prompt on Qwen3.5-4B while keeping
	all other settings at their default values.}
  \label{tab:rollout-ablation}
  \scriptsize
  \resizebox{\textwidth}{!}{\begin{tabular}{cccccccc}
    \toprule
    \textbf{Rollout Number} & \textbf{V* Bench} &
    \textbf{HR-Bench 4K} &
    \textbf{HR-Bench 8K} & \textbf{MMStar} &
    \textbf{MathVista}$_{\boldsymbol{mini}}$ & \textbf{LogicVista} &
    \textbf{Average} \\
    \midrule
    1          & 84.3          & 83.5          & \textbf{81.0} & 74.1          & \textbf{82.9} & 68.9          & 79.1 \\
    \textbf{2} & \textbf{88.5} & 85.0          & 80.5          & \textbf{74.6} & 82.2          & \textbf{69.4} & \textbf{80.0} \\
    4          & 86.4          & \textbf{85.5} & 79.3          & 74.4          & 82.2          & 69.1          & 79.5 \\
    8          & 86.9          & 85.0          & 79.1          & 74.5          & 82.0          & 68.7          & 79.4 \\
    \bottomrule
  \end{tabular}}
\end{table}

\textbf{Number of rollouts.}
Keeping all other settings at their default values, we vary the number of on-policy trajectories per prompt (\cref{tab:rollout-ablation}). All tested rollout counts outperform the baseline on every benchmark. Increasing the count from one to two improves the average by $0.9$ points, while further increases to four or eight yield slightly lower averages. The gains therefore persist across rollout counts, with two rollouts per prompt achieving the highest average score and serving as our default.

\subsection{Effectiveness of Symbolic-to-Visual Correction}
\label{sec:exp-effectiveness}

\mmopd aims to transfer the evidence-use advantage of symbolic conditioning
to image-conditioned reasoning. We examine this transfer in Qwen3.5-4B by
comparing answer-to-evidence attention and the Symbolic Visual Gap before
and after training.

\begin{figure}[t]
	\centering
	\includegraphics[width=0.95\linewidth]{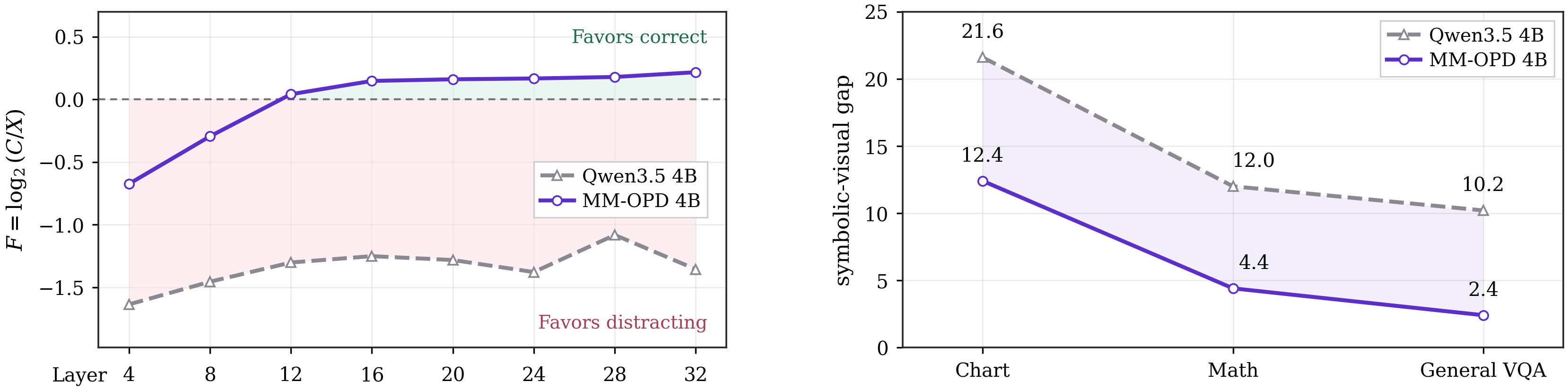}
	\caption{\textbf{Effectiveness of Symbolic-to-Visual Correction.}
		(Left) Layer-wise answer-to-evidence preference
		$F=\log_2(C/X)$ under image conditioning, where positive values favor
		correct evidence. (Right) The Symbolic Visual Gap before and after
		training across three task families.}
	\label{fig:exp-effectiveness}
\end{figure}

\textbf{Evidence preference shifts toward correct evidence.}
We compare image-conditioned answer-to-evidence attention before and
after training using the measure in \cref{sec:analysis-quantitative}.
As shown in \cref{fig:exp-effectiveness} (left), the base model favors
distracting evidence at every observed full-attention layer. In contrast,
\mmopd increases the relative attention to correct evidence across all
observed layers, with correct evidence receiving more attention than
distracting evidence from layer $12$ onward. These results show that
\mmopd reduces the attention bias toward distracting evidence and strengthens
the association between answer generation and the premises supporting
a correct derivation.

\textbf{The Symbolic Visual Gap narrows across tasks.}
We compare image-conditioned performance before and after training
against the base model's symbolic-conditioned accuracy, keeping this
reference fixed for each task family. As shown in
\cref{fig:exp-effectiveness} (right), \mmopd reduces the gap on Chart,
Math, and General VQA. Chart shows the largest absolute reduction, while Math and General
VQA close a larger proportion of their initial gaps. These results show
that \mmopd transfers part of the advantage provided by symbolic views
to image-conditioned performance across all three task families.

Together, the two diagnostics show that \mmopd shifts evidence preference
toward correct evidence while consistently narrowing the performance gap
between image and symbolic conditioning, transferring part of the
symbolic-view advantage to image-only inference.

\section{Related Work}
\label{sec:related}

\textbf{Multimodal Reinforcement Learning.}
Reinforcement learning with verifiable rewards (RLVR)~\citep{guo2025deepseek} has become an important post-train recipe to improving multimodal large language models (MLLMs)~\citep{hong2025glm}. Recent work strengthens visual reasoning by combining supervised cold-start training with RL or scaling RL across diverse multimodal tasks~\citep{huang2026vision,tong2025sketch}. Beyond reasoning over the original visual input, agentic approaches learn to acquire and manipulate visual evidence through operations such as cropping, zooming, and drawing~\citep{hong2026deepeyesv2,su2026pixel,yan2026act}. However, final-answer rewards evaluate the completed response without explicitly specifying which intermediate visual evidence should support it. Our work instead uses token-level self-distillation from symbolic views to guide image-conditioned answer generation.

\textbf{On-Policy Self-Distillation.}
On-policy distillation (OPD) provides dense token-level supervision from a stronger teacher on the student's own rollouts~\citep{agarwal2024policy,lu2025onpolicydistillation}. Evaluating the teacher on prefixes sampled by the current student alleviates the distribution mismatch inherent in off-policy distillation, making OPD a compelling post-training paradigm~\citep{li2026rethinking}.
Recent self-distillation methods obtain this supervision from the same model under more informative conditioning, without requiring a separate teacher model. Self-Distilled Reasoner~\citep{zhao2026self} uses reference solutions, SDPO~\citep{hubotter2026reinforcement} uses textual environment feedback, and Self-Distillation Zero~\citep{he2026self} distills a self-reviser conditioned on the initial response and binary reward. In multimodal settings, Vision-OPD transfers crop-conditioned regional perception to a full-image student~\citep{yuan2026vision}. RLSD combines self-distilled token-level update magnitudes with reward-based update directions~\citep{yang2026self}. Unlike existing multimodal self-distillation methods targeting specific capabilities, such as fine-grained perception in Vision-OPD and mathematical reasoning in RLSD, \mmopd improves a broad range of multimodal abilities through symbolic-to-visual correction.

\textbf{Distillation with Privileged Information.}
Privileged conditioning can improve self-teacher predictions while also introducing preferences that depend on information unavailable to the student. DOPD routes supervision between privileged teacher and student policies according to token-level advantages and probabilities~\citep{yu2026dopd}. Purified OPSD discounts reference-induced preferences and applies the remaining question-conditioned correction to a clean base prediction~\citep{shen2026purified}. \mmopd uses captions and code as privileged symbolic views, applying a residual correction to a clean image-conditioned prediction to improve evidence use.

\section{Conclusion}
In this work, we revisit the assumption of a seamless transition from perception to reasoning. We identify the \emph{Symbolic Visual Gap} and reveal \emph{visual evidence selection} as one more bottleneck between perception and reasoning. Based on the finding that symbolic view elicits the model to assign answer-to-evidence attention to correct visual evidence, we introduce \mmopd, a multimodal on-policy self-distillation framework for \emph{symbolic-to-visual correction} that transfers guidance from symbolic views to the image-conditioned policy through residual token-level targets, steering it toward correct visual evidence. Experiments across benchmarks and model scales demonstrate improvements across a broad range of multimodal abilities rather than a single targeted capability.

\bibliographystyle{plainnat}
\bibliography{references}

\clearpage
\appendix
\section{Additional Case Studies}
\label{sec:appendix-cases}

\begin{figure}[htbp]
  \centering
  \includegraphics[width=\linewidth]{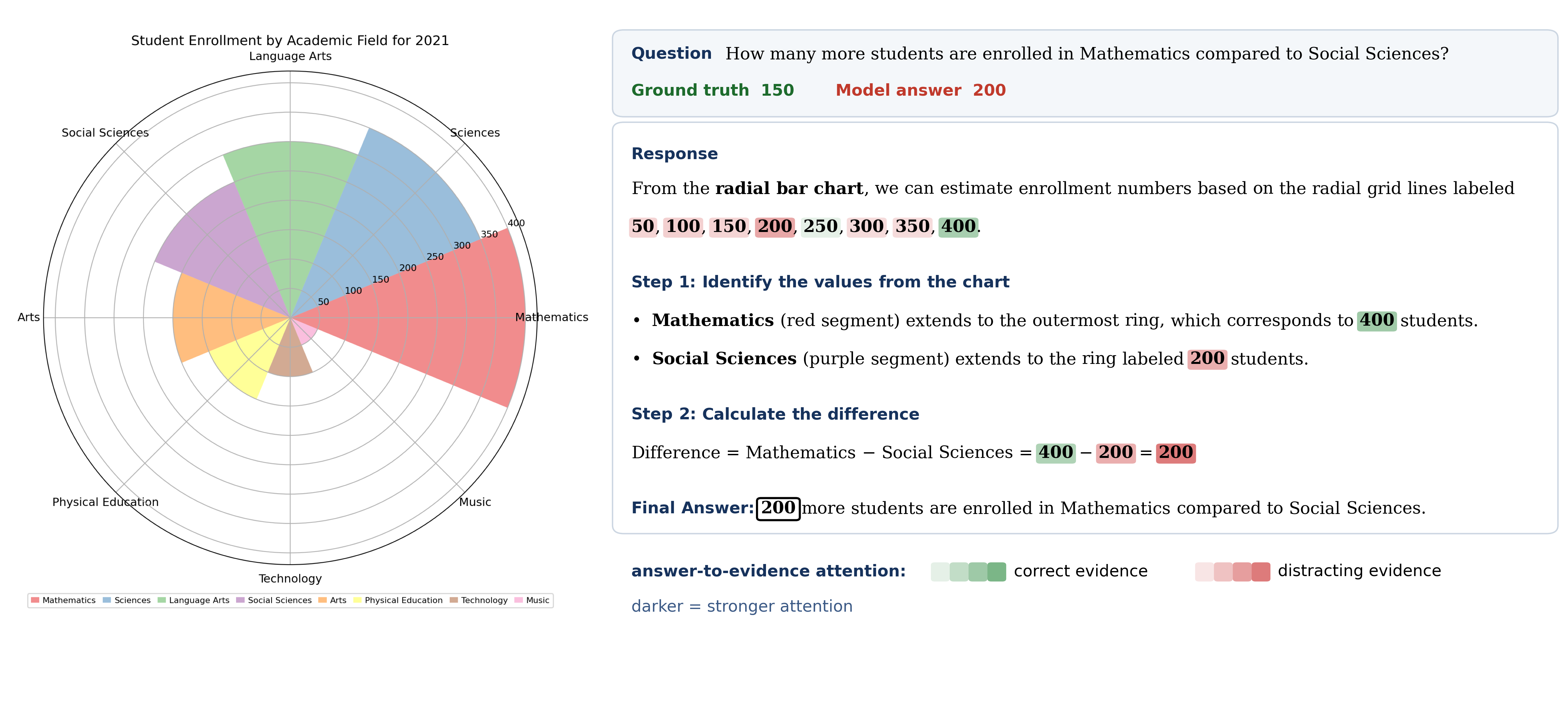}
  \caption{Evidence selection in a radial bar chart. Relevant values are
    available, but distracting evidence is used as the premise for the derivation.}
  \label{fig:sup-case3}
\end{figure}

\begin{figure}[htbp]
  \centering
  \includegraphics[width=\linewidth]{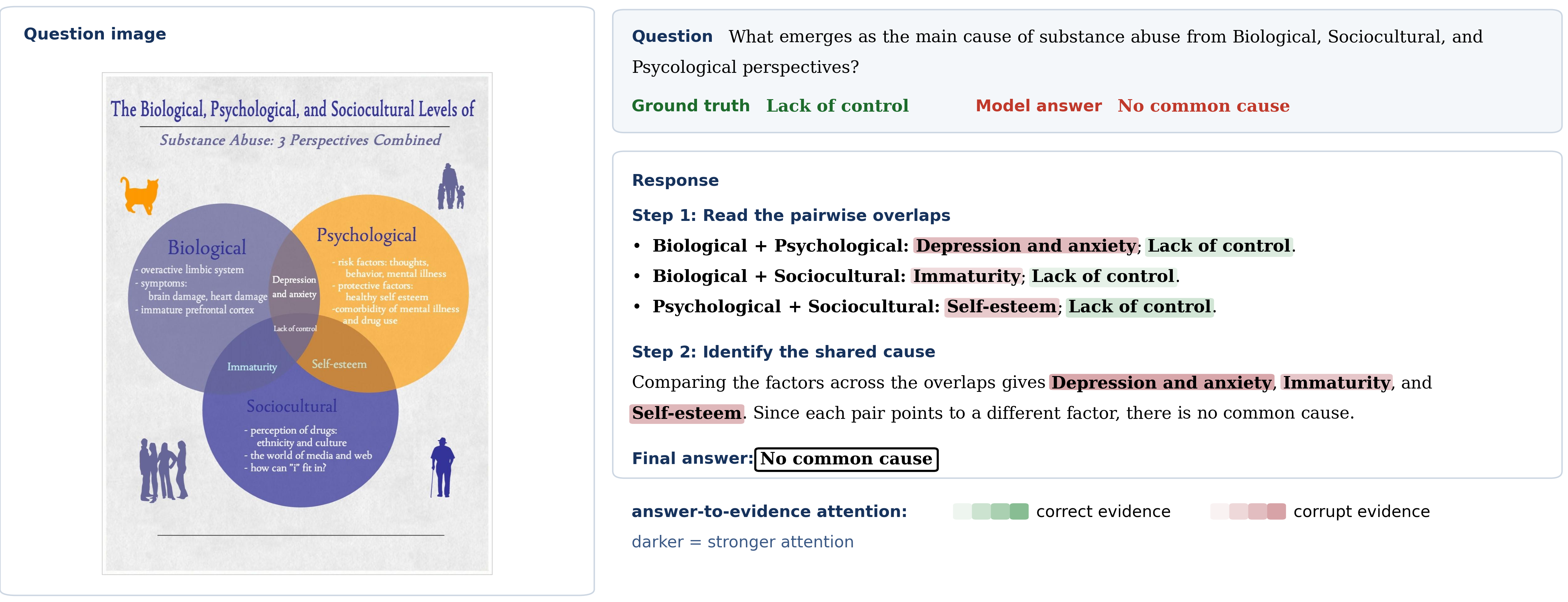}
  \caption{Evidence selection in a Venn diagram. The shared factor is
    verbalized but omitted from the final comparison.}
  \label{fig:sup-case1}
\end{figure}

\begin{figure}[htbp]
  \centering
  \includegraphics[width=\linewidth]{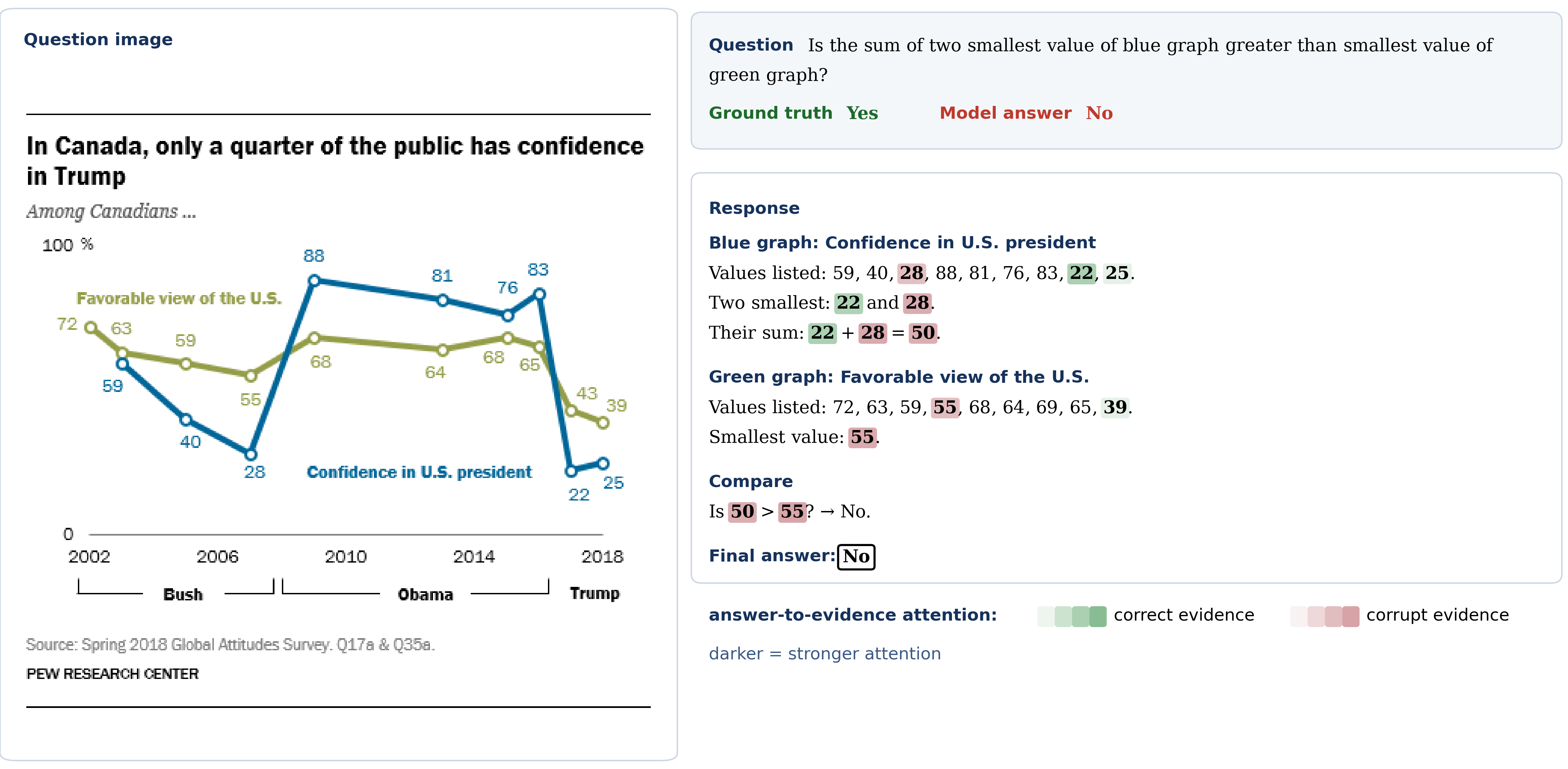}
  \caption{Evidence selection in a line chart. The required values are
    listed, but the computation uses incorrectly selected extrema.}
  \label{fig:sup-case2}
\end{figure}

\Cref{fig:sup-case3,fig:sup-case1,fig:sup-case2} provide additional examples
of visual evidence selection, supplementing \cref{fig:evidence-case} in the
main text.
Green and red mark correct and distracting evidence, respectively, with
darker shading indicating stronger answer-to-evidence attention.

As shown in \cref{fig:sup-case3}, the response contains the relevant chart
values, yet associates a distracting value with the queried category and uses
it in the derivation. The subsequent calculation is valid given that premise,
but yields the wrong answer. Attention is also stronger on the distracting
value used in the derivation. This case further distinguishes the availability
of visual information from its selection as evidence.

A similar pattern appears in \cref{fig:sup-case1}. The response repeatedly
lists ``Lack of control'' across the three overlaps, but compares only
pair-specific factors to conclude that there is no common cause. These
distracting factors receive stronger attention, while the shared factor is
omitted from the comparison. The required information is therefore already
verbalized, but does not guide the final answer.

\clearpage
Likewise, in \cref{fig:sup-case2}, the response lists the required values
$22$, $25$, and $39$, yet selects $22$ and $28$ as the two smallest blue values
and $55$ as the smallest green value. The resulting comparison,
$22+28<55$, yields ``No'' instead of the correct ``Yes'' from $22+25>39$.
Attention also favors the distracting values used in the computation.
Together, these examples support the observation in the main text: having
task-relevant information available in the trace does not ensure that it is
selected as evidence for subsequent reasoning.

\end{document}